\documentclass{article}

\PassOptionsToPackage{numbers, compress}{natbib}

\usepackage[preprint]{nips2023}

\usepackage[utf8]{inputenc} 
\usepackage[T1]{fontenc}    
\usepackage{hyperref}       
\usepackage{url}            
\usepackage{booktabs}       
\usepackage{amsfonts}       
\usepackage{nicefrac}       
\usepackage{microtype}      
\usepackage{xcolor}         
\usepackage{graphicx}
\usepackage{subcaption}

\title{Federated Learning in Temporal Heterogeneity}

\author{%
  Junghwan Lee \\
  H. Milton Stewart School of Industrial and Systems Engineering\\
  Georgia Institute of Technology\\
  \texttt{jlee3541@gatech.edu} \\
}

\begin{document}

\maketitle

\begin{abstract}
In this work, we explored federated learning in temporal heterogeneity across clients. We observed that global model obtained by \texttt{FedAvg} trained with fixed-length sequences shows faster convergence than varying-length sequences. We proposed methods to mitigate temporal heterogeneity for efficient federated learning based on the empirical observation.
\end{abstract}

\section{Introduction}

Federated learning is distributed learning framework where data are used locally (i.e., local devices or local institutions) instead of sharing or storing data through centralized server~\cite{mcmahan2017communication}. Existing studies in federated learning have shown that learning good global model from local models is possible without sharing or storing the entire data. Therefore, federated learning can effectively address privacy concerns related to training deep learning models on large datasets, which gained attention to federated learning.

One of major challenges in federated learning is heterogeneity among local clients. ~\cite{mcmahan2017communication} showed that simply averaging local models often fail to obtain good global model with heterogeneous clients. Numerous studies have been conducted to address the challenge of various types of heterogeneity in federated learning. For example, FedBN~\cite{li2021fedbn} used batch normalization to mitigate heterogeneity of label distribution and FedProx ~\cite{li2020federated} used $l_2$ regularization to control deviance of heterogeneous local models from global model.

The recent success of machine learning in various domains is largely attributed to the use of sequential data containing temporal information. For example, large language models heavily rely on the training on the massive natural language data and disease prediction models require a large number of electronic health records. While sequential data contain innate non-iidness due to varying-length, there have been lack of studies that seek to explore using sequential data. In this paper, we discussed the effect of temporal heterogeneity in sequential data on federated learning. We made the following contributions:

\begin{itemize}
    \item We observed that averaging local models where each local model was trained with the same length of sequences quickly converges than local models with varying-length sequences.

    \item We proposed approaches to mitigate temporal heterogeneity due to varying-length sequences based on the observation.

\end{itemize}

\newpage

\section{Problem Formulation}

We assume $N \in \mathbb{N}$ clients are trained for $E \in \mathbb{N}$ epochs and communicate with global model after $K \in \mathbb{N}$ iterations. Each client $i \in [N]$ has $m_i \in \mathbb{N}$ training examples where each training example consists of $l_j$-length sequence of features in $\mathbb{R}^d$ with label $y_j^i$:  $\{(x_{1,j}^i, ... ,x_{l_j,j}^i) \in \mathbb{R}^{d \times l_j}, (y_{j}^{i}) \in \{0, 1 \}^k  \ : j \in [M], l_j \in \mathbb{N} \}$. Our aim is to train a global model $f^*$ from local models $f_i$ under temporal heterogeneity. Temporal heterogeneity can be defined as how the length of sequences are unequally distributed (e.g., entropy difference between the empirical distribution and uniform distribution).

\section{Experiment}

In this section, we conducted experiments using two datasets: sequential MNIST (sMNIST) and eICU. sMNIST can be considered synthetically manipulated dataset to have temporal heterogenety. eICU dataset has temporal heterogeneity in nature.

\subsection{Setup}

The two most commonly used deep learning architecture for sequential data are Recurrent Neural Networks (RNN) and Transformer~\cite{vaswani2017attention}. We used Gated Recurrent Unit (GRU)~\cite{cho2014learning} and stacked Transformer encoders~\cite{devlin2018bert} for our experiments, which are representative model in each architecture, respectively. We used Adam for optimization with learning rate 0.01.

For federated learning framework, we used FedAvg~\cite{mcmahan2017communication}, the simplest and the most widely used framework in federated learning. We assumed full participation of all clients for all rounds in our experiments. Local training epochs were set to 1 and communication to the global model was conducted after the training of all local models (i.e., full participation) and batch size was set to 32.

\begin{figure}[htbp!]
\centering
\begin{subfigure}[h]{0.45\linewidth}
\includegraphics[width=\linewidth]{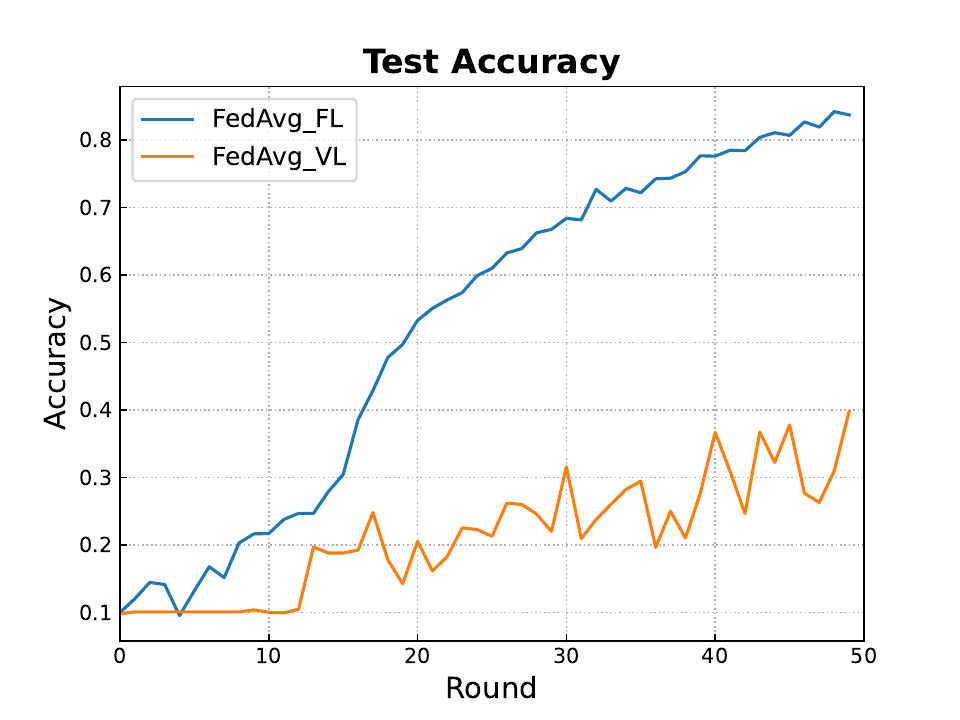}
\end{subfigure}
\begin{subfigure}[h]{0.45\linewidth}
\includegraphics[width=\linewidth]{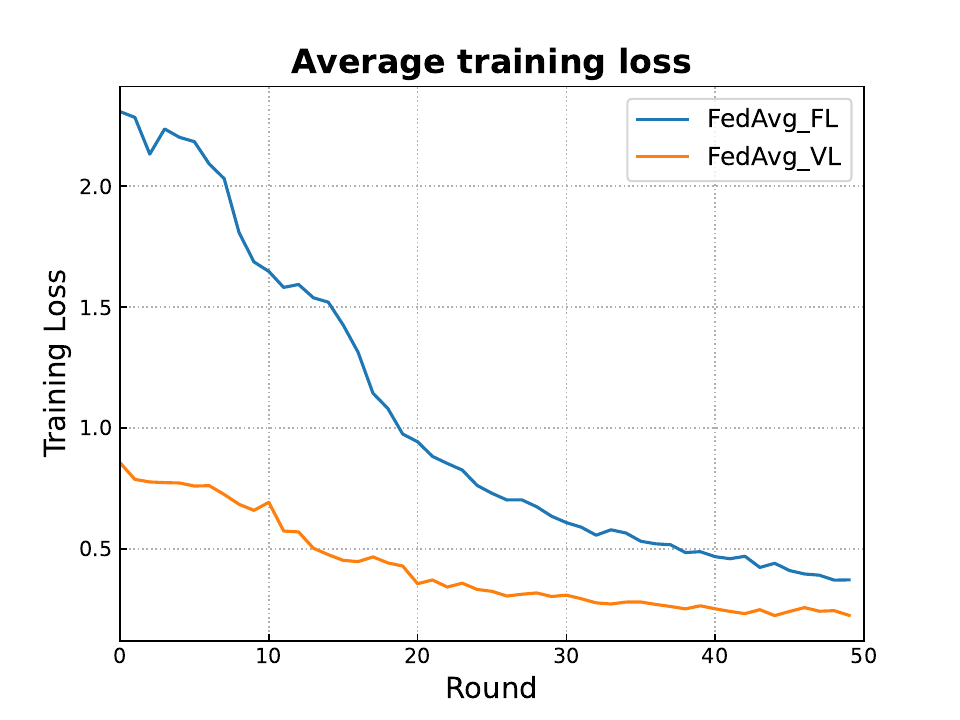}
\end{subfigure}
\caption{(\textbf{Left}) Test accuracy of FedAvg trained with batches of varying-length (FedAvg-VL) and fixed-length (FedAvg-FL). (\textbf{Right}) Average training loss of local models with batches of varying-length (FedAvg-VL) and fixed-length (FedAvg-FL)}
\label{fig:figure1}
\end{figure}

\subsection{Sequential MNIST}

Sequential MNIST (sMNIST) is used to learn long-term dependency~\cite{le2015simple}. In sMNIST, original MNIST images were flattened in scanline order (i.e., starting from the top left to the bottom right) where 784 pixels are represented in a single sequence. We constructed two different datasets to explore the effect of temporal heterogeneity. 

For the first dataset, we divided MNIST into five subsets where each subset has the same label distribution and the number of images. Then the images in each subset were randomly re-sized into one of $[14 \times 14, 17 \times 17, 21 \times 21, 24 \times 24, 28 \times 28]$. This dataset contains both temporal heterogeneity within clients (varying-length of sequences in a sigle client) and between clients (different distribution of sequence lengths). We refer this dataset as varying-length dataset (VL dataset). 

The second dataset also has the five subsets showing the same label distribution and the number of images, but all images in the same subset were re-sized into one of $[14 \times 14, 17 \times 17, 21 \times 21, 24 \times 24, 28 \times 28]$. While temporal heterogeneity exists both inside and across clients in the first dataset, temporal heterogeneity only exists across clients in the second dataset. We refer the second dataset as fixed-length dataset (FL dataset).

\begin{figure}[htbp!]
\centering
\begin{subfigure}[h]{0.45\linewidth}
\includegraphics[width=\linewidth]{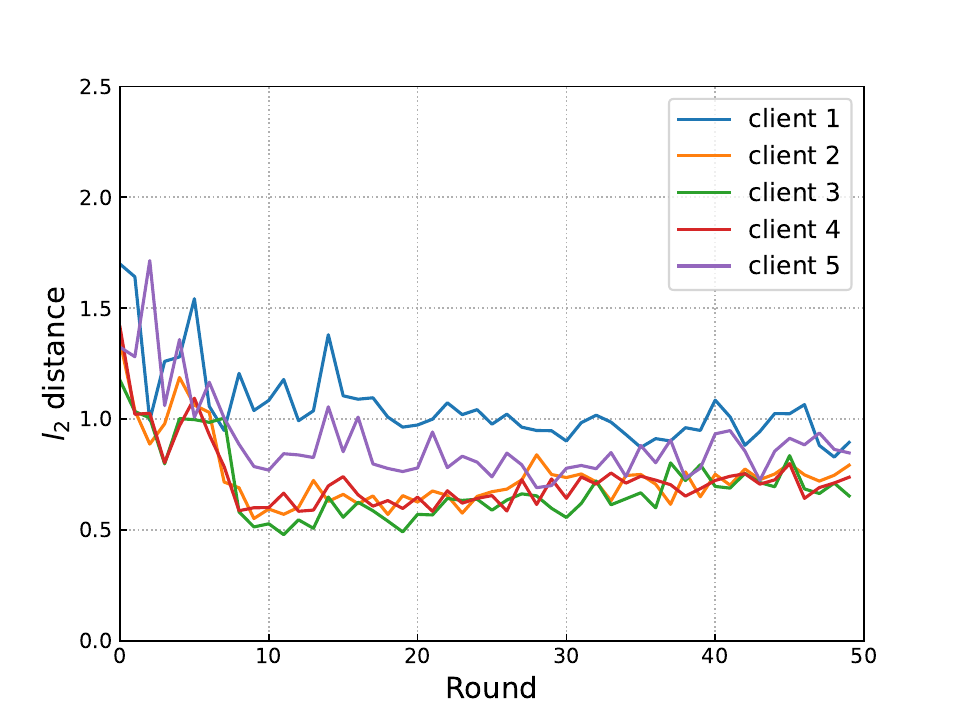}
\end{subfigure}
\begin{subfigure}[h]{0.45\linewidth}
\includegraphics[width=\linewidth]{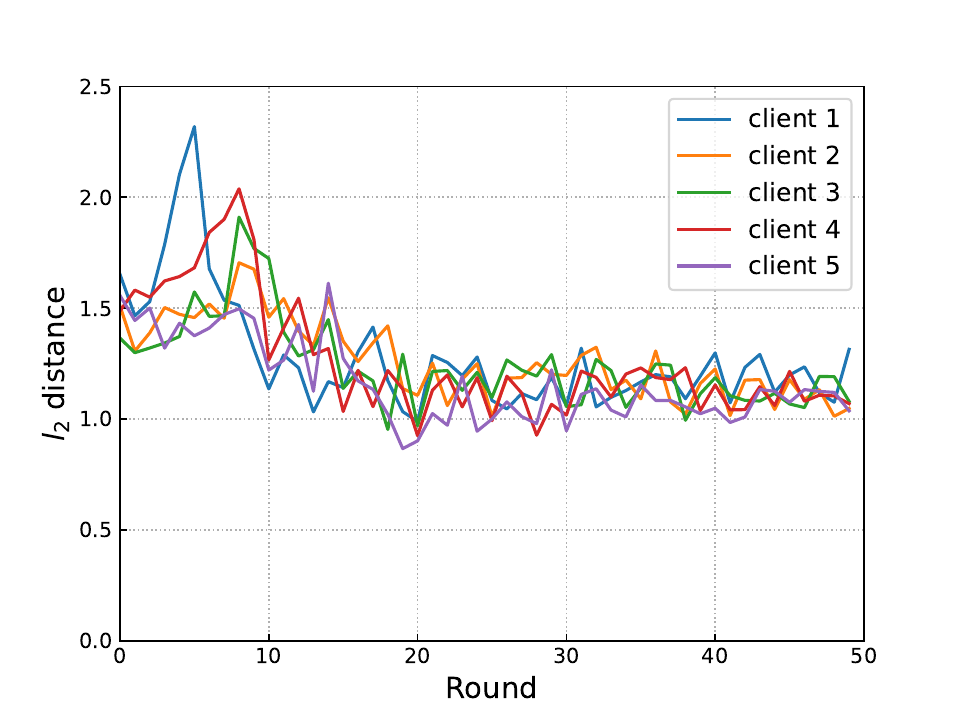}
\end{subfigure}
\caption{(Left) Test accuracy of FedAvg trained with batches of varying-length (FedAvg-VL) and fixed-length (FedAvg-FL). (Right) Average training loss of local models with batches of varying-length (FedAvg-VL) and fixed-length (FedAvg-FL)}
\label{fig:figure2}
\end{figure}

Figure~\ref{fig:figure1} shows test accuracy and average training loss per communication round on VL dataset and FL dataset. We observed that FedAvg on FL dataset quickly converges than VL dataset. Figure~\ref{fig:figure2} shows $l_2$-distance between each local model and global model in FedAvg on the two datasets. While FedAvg on VL dataset and FL dataset both shows convergence to the global model, FedAvg on FL dataset shows different behavior from FedAvg on VL dataset. As local model divergence behavior has crucial impact on the performance of global model especially in non-iid setting~\cite{zhu2021federated,li2020federated}, this warrants further investigation.

\subsection{Intensive Care Unit Acuity Prediction}

This experiment will be added after theoretical justification based on the empirical observation using sMNIST.

\section{Proposed Method}

\begin{figure}[htbp!]
\centering
\begin{subfigure}[h]{0.6\linewidth}
\includegraphics[width=\linewidth]{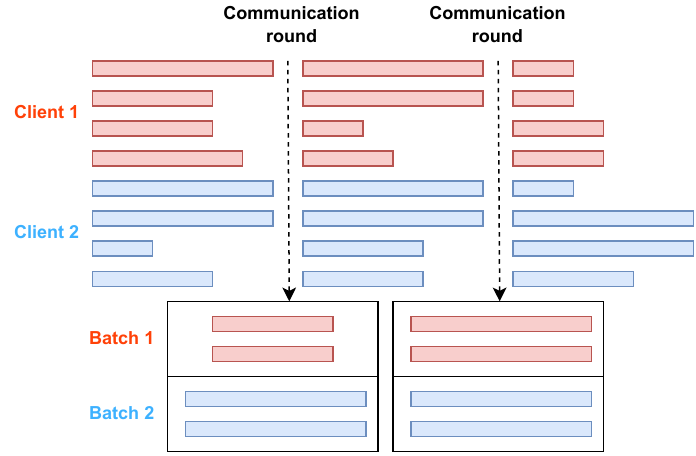}
\end{subfigure}
\caption{Visual description of batch alignment. Color represents client membership of data. Solid box represents local training batch. Dashed lines represent communication round.}
\label{fig:figure3}
\end{figure}

\subsection{Batch Alignment}

Figure~\ref{fig:figure3} depicts batch alignment for federated learning. As we observed that training with batches containing fixed-length sequences, batch alignment helps efficient convergence of global model by aligning the length of sequences in a batch for each local model.

\subsection{Normalization}

Previous research has shown that proper normalization is beneficial for global model convergence ~\cite{yang2022towards, li2021fedbn}. However, application of normalization to recurrent neural networks or Transformer requires careful consideration due to various reasons (e.g., gradient flow)~\cite{cooijmans2016recurrent, shen2020powernorm, laurent2016batch}. Further investigation requires to justify the benefits of sequential normalization over normalization of the learned representation for the entire sequence.

\section{Future Works}

We admit that we did not conducted all planned experiments due to time limit. This works are purely based on empirical observation, therefore theoretical justification will be required. In order for theoretical analysis, we need clear mathematical definition of temporal heterogeneity. Neural Tangent Kernel~\cite{jacot2018neural} can be used for theoretical analysis (motivated by \cite{li2021fedbn}), which can be combined with empirical observation of the tracking of gradient w.r.t. the varying length of sequences.

Additionally, we did not include the results using Transformer encoder, which is the most commonly used deep learning model for sequential data. Therefore, future works will also include the same experiments using Transformer encoder.

\vspace{-0.1in}

{
\small
\bibliography{main}
}


\end{document}